\documentclass[mlabstract]{jmlr_camera}

\usepackage[hpos=300px,vpos=70px]{draftwatermark}
\SetWatermarkText{\test}
\SetWatermarkScale{1}
\SetWatermarkAngle{0}

\usepackage{longtable}

\usepackage{booktabs}
\usepackage[load-configurations=version-1]{siunitx} 

\usepackage{wrapfig}
\usepackage{caption}
\usepackage{subcaption}
\usepackage{graphicx}

\theorembodyfont{\upshape}
\theoremheaderfont{\scshape}
\theorempostheader{:}
\theoremsep{\newline}

\newcommand{\ourmethod}{\textsc{GMAN}}
\newcommand{\featuregroupgnan}{\textsc{ExtGNAN}}
\newcommand{\h}{\mathbf{h}}
\jmlrvolume{}
\firstpageno{1}

\jmlryear{2025}
\jmlrworkshop{Symmetry and Geometry in Neural Representations}

\title[Short Title]{Graph Mixing Additive Networks}

  \author{\Name{Maya Bechler-Speicher} \Email{mayabs@meta.com}\\
  \addr Meta
  \AND
  \Name{Andrea Zerio} \Email{anze@dcm.aau.dk}\\
  \addr {Center of Excellence for Molecular Prediction of IBD, PREDICT,
Department of Clinical Medicine, Aalborg University}
  \AND
  \Name{Maor Huri} \Email{maorhury@mail.tau.ac.il}\\
  \addr {Sagol School of Neuroscience
Tel Aviv University}
  \AND
  \Name{Marie Vibeke Vestergaard} \Email{pepijnrh@kth.se}\\
  \addr {Center of Excellence for Molecular Prediction of IBD, PREDICT,
Department of Clinical Medicine, Aalborg University}
  \AND
  \Name{Ran Gilad-Bachrach} \Email{rgb@tauex.tau.ac.il}\\
  \addr {Department of Bio-Medical Engineering and Edmond J. Safra Center for Bioinformatics, Tel-Aviv University}
  \AND
  \Name{Tine Jess} \Email{jess@dcm.aau.dk }\\
  \addr {Center of Excellence for Molecular Prediction of IBD, PREDICT,
Department of Clinical Medicine, Aalborg University\\
Department of Gastroenterology and Hepatology, Aalborg University Hospital}
  \AND
\Name{Samir Bhatt} \Email{ samir.bhatt@sund.ku.dk}\\
  \addr {University of Copenhagen\\
  Imperial College London}
  \AND
      \Name{Aleksejs Sazonovs} \Email{alesaz@dcm.aau.dk}\\
  \addr {Center of Excellence for Molecular Prediction of IBD, PREDICT,
Department of Clinical Medicine, Aalborg University}
 }

\begin{document}

\maketitle

\begin{abstract}
We introduce \ourmethod{}, a flexible, interpretable, and expressive framework that extends Graph Neural Additive Networks (GNANs) to learn from sets of sparse time-series data. \ourmethod{} represents each time-dependent trajectory as a directed graph and applies an enriched, more expressive GNAN to each graph. It allows users to control the interpretability-expressivity trade-off by grouping features and graphs to encode priors, and it provides feature, node, and graph-level interpretability. On real-world datasets, including mortality prediction from blood tests and fake-news detection, \ourmethod{} outperforms strong non-interpretable black-box baselines while delivering actionable, domain-aligned explanations.
\end{abstract}


\section{Introduction}
\label{sec:intro}

Many datasets contain heterogeneous measurements sampled at irregular and asynchronous times. For example in clinical data, A patient record, for example, may comprise multiple medical tests taken at task-dependent frequencies, which can be naturally viewed as a set of sparse temporal trajectories. As another example, news propagation unfolds along platform-specific trajectories through a social network, with timing and topology jointly shaping diffusion. Conventional pipelines regularize irregular data via imputation or grid alignment, using recurrent, attention, or diffusion-based imputers \citep{cao2018brits, tashiro2021csdi, wu2022timesnet, du2023saits}. However, such preprocessing can distort dynamics, and obscure conditional dependencies.

To address this, we introduce Graph Mixing Additive Networks (\ourmethod{}), an interpretable framework for learning from sets of temporally sparse graphs. Concretely, \ourmethod{} treats each trajectory as a directed path graph (with edges parameterized by elapsed time) and each sample as a set of such graphs, preserving temporal distances and permutation invariance across channels/paths. Methodologically, \ourmethod{} (i) encodes each trajectory with an Extended GNAN (ExtGNAN) \cite{bechler2024intelligible} module that aggregates over temporal edges and applies univariate or multivariate shape functions to feature groups; and (ii) non-linearly combines trajectories via a partition of the set into subsets, enabling a tunable trade-off between interpretability and expressivity.

 We evaluate \ourmethod{} on two structurally distinct problems central to learning representations for irregular signals: (1) In-hospital mortality prediction from routine blood-test trajectories, and (2) fake-news detection from social propagation paths (GossipCop). \ourmethod{} achieves state-of-the art performance while providing fine-grained temporal and graph-level attributions. Finally we prove that GMAN is strictly more expressive than GNAN, and that grouped (non-singleton) subset mixing strictly increases expressivity over singleton subsets.

\section{Method}

In this section, we present \ourmethod{}, an interpretable and flexible method for effectively learning over sets of trajectories of varying size. Let $S = \{G_1, \dots, G_m\}$ be a \emph{set} of $m$ graphs. Each node $v$ is associated with a feature vector $x_v \in \mathbb{R}^d$ and a time-stamp $t_{v}$.
For instance, in the case of a patient's blood test data, each graph corresponds to a specific biomarker, and each node within the graph represents an individual measurement of that biomarker, annotated with its feature vector and time of collection.
We denote the entry $c$ of a vector $\h$ by $[\h]_c$, and the set of entries corresponding to a set of features $S$ by $[\h]_S$.

We assume that the graphs in $S$ are partitioned into $k, 1\leq k\leq m$ disjoint subsets $S_k$ such that $\bigcup_{i=1}^k S_i = S$.
The partition $\{S_i\}_{i=1}^k$ provides a flexible trade-off between expressivity and interpretability.
\ourmethod{} linearly aggregates representations of the subsets of $S$ to form a final set representation, and then assigns a single label to $S$.
The level of interpretability that \ourmethod{} provides for each graph depends on the size of the subset it belongs to. When a subset contains a single graph, \ourmethod{} offers fine-grained, node-level importance scores. In contrast, for larger subsets, it provides only set-level importance scores—trading interpretability for improved expressivity. This design enables a flexible trade-off between interpretability and expressive power, controlled by the chosen partitioning strategy.

$$
\h_i = \Phi_i(S_i),
$$

Then, it produce a representation for the whole set, $\h_S$ by summing the subsets' $\h_S = \sum_{i=1}^k h_i$. 
Finally, to produce the label, it sums over the $d$ entries of $\h_S$. Overall:

\begin{equation}\label{eq:gman}
    \ourmethod(S) = \sum_{c=1}^d \sum_{i=1}^k [\Phi_i(S_i)]_c
\end{equation}

Where $\Phi_i(S_i) = \h_{S_i}$ is a representation of the subset $S_i$.

For subsets of size one, $\Phi_i(S_i)$ applies an Extended GNAN (ExtGNAN), as described in Section \ref{sec:ExtGNAN}. For subsets containing multiple graphs, a ExtGNAN is applied to each graph, followed by a DeepSet aggregation~\citep{zaheer2018deepsets} over the resulting vectors. Importantly, each subset is assigned its own ExtGNAN, and all graphs within a subset share the same one.
A DeepSet first applies a neural network $f: \mathbb{R}^d \rightarrow \mathbb{R}^d$ for each vector in the set $\{h_l\}_{G_l\in S_i}$, sums the results, and then applies another neural network $g:  \mathbb{R}^d \rightarrow \mathbb{R}^d$.

$$
g\left(\sum_{i\in S_2} f(h_i)\right)
$$

Here, $g$ and $f$ are neural networks of arbitrary depth and width.
We now turn to define \featuregroupgnan.

\subsection{ExtGNAN} \label{sec:ExtGNAN}
In GNAN \cite{bechler2024intelligible}, univariate neural networks are applied to each feature of each node in isolation, to learn a representation for a graph. 
This has the benefit of generating interpretable models as features do not mix non-linearly. Nonetheless, when interactions between features are crucial for the task, it may result in sub-par performance. Therefore, \featuregroupgnan{} extends GNAN by allowing multivariate neural networks to operate on groups of features to gain accuracy at the cost of reducing the feature-level interpretability of the model. 

Assume that the features are partitioned into \( K \) subsets $\{F_l\}_{l=1}^K$. For any subset of features greater than one, \featuregroupgnan{} applies a multivariate neural network for all the features in the subset together, instead of a univariate neural network for each one separately. 
To learn a representation of a graph $G$, \featuregroupgnan{} first computes representations for the nodes of $G$ as follows. 

\featuregroupgnan{} learns a distance function $\rho(x;\theta): \mathbb{R}\rightarrow  \mathbb{R}$ and a set of feature shape functions 
 $\{\psi_l\}_{l=1}^K,  \psi_l(X;\theta_k): \mathbb{R}^{|F_l|} \rightarrow \mathbb{R}^{|F_l|}$.
 Each of these functions is a neural network of arbitrary depth. For brevity, we omit the parameterization $\theta$ and $\theta_k$ for the remainder of this section.

The entries of the representation of node \( j \) corresponding to the indices of the features in $F_l$, denoted as \( [\mathbf{h}_j]_{F_l} \), is computed by summing the contributions of the features in the subset \( F_l \) from all nodes in the graph:
\[
[\mathbf{h}_j]_{F_l} = \sum_{w \in V}\rho\left(\Delta(w, j)\right) \cdot \psi_l\left([\mathbf{X}_w]_{F_l}\right),
\]
where $\Delta(w, j) = t_w - t_j$ and $[\mathbf{X}_w]_{F_l}$ are the features of node $w$ corresponding to the subset $F_l$.

Overall, the full representation of node \( j \) can be written as:
\[
\mathbf{h}_j = \big([\mathbf{h}_j]_{F_1}, [\mathbf{h}_j]_{F_2}, \dots, [\mathbf{h}_j]_{F_K}\big).
\]

Then \featuregroupgnan{} produces a graph representation by summing the node representations,
\begin{equation}\label{eq:graph_rep_ext}
    \mathbf{h}_G = \sum_{i \in V} \mathbf{h}_i.
\end{equation}

\subsection{Expressivity}
In this section we provide two theoretical results of the expressivity of \ourmethod{}. Proofs are in the Appendix.

\begin{theorem}
    \ourmethod{} is strictly more expressive than GNAN.
\end{theorem}

\begin{theorem}
    Let $S$ be a set of graphs $\{G_i\}_{j=1}^m$. Let   $S_1 = \{S_i\}_{i=1}^m$ be a partition of $S$ such that $|S_i|=1$. Let $S_2 = \{S_i\}_{i=1}^k$ such that there exists $k$ with $|S_k|>1$. 
    with a subset partition $\{S_i\}_{i=1}^k$. Then a \ourmethod{} trained over $S_2$ is strictly more expressive than a \ourmethod{} trained over $S_1$. 
\end{theorem} 

\section{Empirical Evaluation}

We present preliminary results evaluating \ourmethod{} on two diverse tasks: (1) predicting mortality from biomarker (blood tests) trajectories, and (2) detecting fake news propagation patterns in social networks. These tasks vary significantly in domain and graph structure, allowing us to assess generalization and interpretability in real-world settings. Full experimental setup and dataset details are detailed in the Appendix.
\begin{wraptable}{r}{0.4\textwidth}
\vspace{-6pt}
\centering
\caption{Evaluation of \ourmethod{} in-hospital mortality prediction (P12).}
\label{tab:p12_perf}

\footnotesize
\begin{tabular}{@{}lc@{}}
\toprule
Method & AUROC \\
\midrule
Transformer & 65.1 ± 5.6 \\
Trans-mean & 66.8 $\pm$ 4.2 \\
GRU-D      & 67.2 $\pm$ 3.6 \\
SeFT       & 66.8 $\pm$ 0.8 \\
mTAND      & 65.3 $\pm$ 1.7 \\
IP-Net     & 72.5 $\pm$ 2.4 \\
DGM$^2$    & 71.2 $\pm$ 2.5 \\
MTGNN      & 67.5 $\pm$ 3.1 \\
RAINDROP   & 72.1 $\pm$ 1.3 \\
\midrule
\ourmethod & \textbf{76.64 $\pm$ 1.2} \\
\bottomrule
\end{tabular}

\end{wraptable}

\paragraph{In-hospital mortality prediction}
We use the PhysioNet2012 (P12) dataset, introduced by \cite{goldberger2000physiobank}, which contains records from 11,988 ICU patients, following the exclusion of 12 samples deemed inappropriate according to the criteria in Horn et al. (2020). For each patient, time series measurements from 36 physiological signals (excluding weight) were recorded over the initial 48 hours of ICU admission. Additionally, each patient has a static profile comprising 9 features, including demographic and clinical attributes such as age and gender. The dataset is labelled for a binary classification task: predicting in-hospital mortality.

We compare \ourmethod{} to other $9$ baselines evaluated in \cite{zhang2022graphguidednetworkirregularlysampled} and use their splits,
including: \textit{Transformer}~\citep{vaswani2017attention}, \textit{Trans-mean}, \textit{GRU-D}~\citep{che2016recurrentneuralnetworksmultivariate}, \textit{SeFT}~\citep{horn2020setfunctionstimeseries},  \textit{mTAND}~\citep{shukla2021multitimeattentionnetworksirregularly}, IP-Net\citep{shukla2019interpolationpredictionnetworksirregularlysampled}, DGM$^2$\citep{wu2021dynamicgaussianmixturebased} and MTGNN~\citep{wu2020connectingdotsmultivariatetime}. 

We conducted a grid search by training on the training set and evaluating on the validation set. We selected the best performing model over the validation set. We report the average AUROC score and standard-deviation of the selected configuration with $3$ seeds.

The complete subsets information is provided in the Appendix.

\begin{wraptable}{R}{0.4\textwidth}

\centering
\caption{Evaluation of \ourmethod{} on fake news detection (GOS).}
\label{tab:fake_news_result}

\footnotesize
\begin{tabular}{lc}

\toprule
\textbf{Methods} & \textbf{Accuracy} \\
\midrule
GATv2 & 96.10 ± 0.3\\
GraphConv &  96.77 ± 0.1\\
GraphSage &  94.45 ± 1.5\\
GCNFN &  96.52 ± 0.2 \\

\midrule
\ourmethod & \textbf{ 97.34 ± 0.2} \\
\bottomrule
\end{tabular}

\end{wraptable}

\paragraph{Fake News Detection.}
The GossipCop (GOS) dataset contains 5,464 retweet cascades, each represented as a tree-structured graph. We decompose cascades into directed propagation paths and represent each as a graph. Graphs are grouped into subsets based on structure, enabling controlled non-linear aggregation. More details are provided in the Appendix. Results are presented in Table~\ref{tab:fake_news_result}, showing \ourmethod{} outperforms all GNN baselines, including GATv2 \citep{brody2022attentivegraphattentionnetworks}, GraphConv \citep{morris2021weisfeiler}, GraphSAGE \citep{hamilton2018inductiverepresentationlearninglarge} and GCNFN \citep{monti2019fakenewsdetectionsocial}.

\section{Conclusion}
In this extended abstract we introduced \ourmethod{}, a work-in-progress framework for learning from sets of sparse temporal graphs that achieves strong predictive performance while preserving multi-resolution interpretability. By combining distance-aware additive encoders with a flexible subset-mixing mechanism, \ourmethod{} enables fine-grained attributions when needed, and non-linear modelling when beneficial. Together, these features support actionable insight in high-stakes domains such as healthcare, while maintaining the flexibility to generalize across diverse data modalities.
\newpage
\bibliography{pmlr-sample}

\newpage

\appendix

\section{Node, graph and set importance}
\ourmethod{} retains all interpretability properties of GNAN \citep{bechler2024intelligible}, including feature-level and node-level importance. However, it extends beyond GNAN by operating on sets of graphs rather than single graphs, enabling additional forms of interpretability such as graph-level and subset-level importance. Because \ourmethod{} allows a flexible trade-off between interpretability and expressivity, permitting non-linear mixing within graph subsets, some adaptations are required to extract meaningful attributions under this more expressive regime. We can extract the total contribution of each node \( j \) to the prediction by summing the contributions of the node across all feature sets. This is only valid when the node belongs to a graph that is not combined non-linearly with other graphs, i.e., it belongs to a subset of size one. \\

\noindent
Therefore, the contribution of node $j$ is
 \begin{equation}\label{eq:node_importance}
      \text{TotalContribution}(j) =  \sum_{l=1}^K [\mathbf{h}_j]_{F_k}=  
      \sum_{w \in V}\rho\left(\Delta(w, j)\right) \sum_{l=1}^K \psi_k\left([\mathbf{x}_w]_l, l\in F_k\right).
 \end{equation}

\noindent The contribution of a graph $G$ is then

$$
  \text{TotalContribution}(G) = \sum_{v\in G} \text{TotalContribution}(v).
$$

\noindent For graphs that are mixed non-linearly, i.e., graphs that belong in subsets of size greater than one, interpretability is more limited, and we can only provide the total contribution of the set to the final prediction
\begin{equation}\label{eq:set_importance}
    \text{TotalContribution}(S) = \sum_{l=1}^K [\mathbf{S}]_{F_k}.
\end{equation}

\section{Expressivity properties}

In this section we provide a theoretical analysis of the expressivness of \ourmethod{}.

\begin{theorem}
    \ourmethod{} is strictly more expressive than GNAN.
\end{theorem}

\paragraph{Proof of Theorem 1}
We will prove that GMAN is strictly more expressive than GNAN.
To prove this, we use a ground truth function that is a feature-level XOR.
Let a single-node graph be endowed with binary features 
\(x=(x_1,x_2)\in\{0,1\}^2\) and define the target  
\(f_{\oplus}(x)=x_1\oplus x_2\).

First we will show that GNAN cannot express \(f_{\oplus}\).
A GNAN scores the graph by
\(\hat{y}=\sigma\!\bigl(\phi_1(x_1)+\phi_2(x_2)\bigr)\),
where each \(\phi_i\) is univariate.  
Put \(a=\phi_1(0),\,b=\phi_1(1),\,c=\phi_2(0),\,d=\phi_2(1)\).
To match the XOR truth-table there must exist a threshold \(\tau\) such that  
\[
a+c<\tau,\quad b+c>\tau,\quad a+d>\tau,\quad b+d<\tau.
\]
Summing the first and last inequalities yields \(a+b+c+d<2\tau\),  
while the middle pair gives \(a+b+c+d>2\tau\)---a contradiction.  
Thus no GNAN realises \(f_{\oplus}\).

Now we will show that \ourmethod{} can express \(f_{\oplus}\).
Place the two features in the same subset \(F=\{x_1,x_2\}\) and choose the
subset-network  
\[
\psi_F(x_1,x_2)=x_1+x_2-2x_1x_2.
\]
For the four binary inputs this mapping returns \((0,1,1,0)\), exactly \(f_{\oplus}\).  
Hence  \ourmethod{} represents a function unattainable by GNAN, proving that  \ourmethod{} is strictly 
more expressive.\\

\begin{theorem}
    Let $S$ be a set of graphs $\{G_i\}_{j=1}^m$. Let   $S_1 = \{S_i\}_{i=1}^m$ be a partition of $S$ such that $|S_i|=1$. Let $S_2 = \{S_i\}_{i=1}^k$ such that there exists $k$ with $|S_k|>1$. 
    with a subset partition $\{S_i\}_{i=1}^k$. Then a \ourmethod{} trained over $S_2$ is strictly more expressive than a \ourmethod{} trained over $S_1$. 
\end{theorem} 

\paragraph{Proof of Theorem 2}

 Let $S$ be a set of graphs $\{G_i\}_{j=1}^m$. Let   $S_1 = \{S_i\}_{i=1}^m$ be a partition of $S$ such that $|S_i|=1$. Let $S_2 = \{S_i\}_{i=1}^k$ such that there exists $k$ with $|S_k|>1$. 
    with a subset partition $\{S_i\}_{i=1}^k$. We will prove that a \ourmethod{} trained over $S_2$ is strictly more expressive than a \ourmethod{} trained over $S_1$. 

To prove this, we use a ground truth function that is a set-level XOR.
Let every graph $G_i$ carry a single binary feature $x_i\in\{0,1\}$ and let the
\textsf{ExtGNAN} encoder return this feature unchanged, i.e.\ $h(G_i)=x_i$.
Denote a set containing two graphs by $S=\{G_1,G_2\}$ and define the
permutation-invariant target
\[
f_\oplus(S)=x_1\oplus x_2 .
\]

\noindent \underline{Singleton partition ($S_1$).}
If each graph is placed in its own subset, GMAN aggregates \emph{additively}: the
model output is
\[
\hat y \;=\;\phi(x_1)+\phi(x_2),
\]
because the final GMAN stage simply sums subset scores.
Write $a=\phi(0)$ and $b=\phi(1)$.  To realise $f_\oplus$ via a threshold
$\tau$ we would need
\[
a+a<\tau,\quad b+a>\tau,\quad a+b>\tau,\quad b+b<\tau.
\]
Adding the first and last inequalities yields $a+b<\tau$, while the middle
pair gives $a+b>\tau$—a contradiction.  Hence GMAN$_{S_1}$ cannot represent $\,f_\oplus$. \\

\noindent \underline{Paired partition ($S_2$)}
Group the two graphs together and use a DeepSet
$\Phi(S_2)=g\!\bigl(\sum_{i=1}^2 f(x_i)\bigr)$ with
$f(x)=x$ and $g(s)=s(2-s)$.  Then
\[
g\bigl(x_1+x_2\bigr)=
\begin{cases}
0 & (x_1,x_2)=(0,0)\text{ or }(1,1),\\
1 & (x_1,x_2)=(0,1)\text{ or }(1,0),
\end{cases}
\]
exactly $f_\oplus$.  The final GMAN sum over feature channels leaves this
value unchanged, so GMAN$_{S_2}$ realises $\,f_\oplus$.\\

\noindent \underline{Strict separation.}
Because $f_\oplus$ is representable by GMAN$_{S_2}$ but not by GMAN$_{S_1}$, the former is strictly more expressive.




\section{Dataset Details}
\label{app:datasets}

\subsection{PhysioNet P12}
\label{app:p12}
We provide the full list of the 36 physiological signals and 3 static patient features used in our experiments.

\begin{enumerate}
    \item Alkaline phosphatase (ALP): A liver- and bone-derived enzyme; elevations suggest cholestasis, bone disease, or hepatic injury. \\

    \item Alanine transaminase (ALT): Hepatocellular enzyme; increased values mark acute or chronic liver cell damage. \\

    \item Aspartate transaminase (AST): Enzyme in liver, heart, and muscle; rises indicate hepatocellular or muscular injury. \\

    \item Albumin: Major plasma protein maintaining oncotic pressure and transport; low levels reflect inflammation, malnutrition, or liver dysfunction. \\

    \item Blood urea nitrogen (BUN): End-product of protein catabolism cleared by the kidneys; elevation signals renal impairment or high catabolic state. \\

    \item Bilirubin: Hemoglobin breakdown product processed by the liver; accumulation indicates hepatobiliary disease or hemolysis. \\

    \item Cholesterol: Circulating lipid essential for membranes and hormones; dysregulation is linked to cardiovascular risk. \\

    \item Creatinine: Waste from muscle metabolism filtered by the kidneys; higher levels imply reduced glomerular filtration. \\

    \item Invasive diastolic arterial blood pressure (DiasABP): Pressure during ventricular relaxation; low readings may reflect vasodilation or hypovolemia. \\

    \item Fraction of inspired oxygen (FiO$_2$): Proportion of oxygen delivered; values above ambient air denote supplemental therapy. \\

    \item Glasgow Coma Score (GCS): Composite neurologic score for eye, verbal, and motor responses; scores $\leq8$ indicate severe impairment. \\

    \item Glucose: Principal blood sugar; hypo- or hyper-glycemia can cause neurologic compromise and metabolic instability. \\

    \item Serum bicarbonate (HCO$_3$): Key extracellular buffer; low levels signal metabolic acidosis, high levels metabolic alkalosis or compensation. \\

    \item Hematocrit (HCT): Percentage of blood volume occupied by red cells; reduced values denote anemia, elevated values hemoconcentration. \\

    \item Heart rate (HR): Beats per minute reflecting cardiac demand; tachycardia indicates stress or shock, bradycardia conduction disorders. \\

    \item Serum potassium (K): Crucial intracellular cation; deviations predispose to dangerous arrhythmias. \\

    \item Lactate: By-product of anaerobic metabolism; elevation marks tissue hypoxia and shock severity. \\

    \item Invasive mean arterial blood pressure (MAP): Time-weighted average arterial pressure; low values threaten organ perfusion. \\

    \item Mechanical ventilation flag (MechVent): Binary indicator of ventilatory support; presence denotes respiratory failure or peri-operative care. \\

    \item Serum magnesium (Mg): Cofactor for numerous enzymatic reactions; abnormalities contribute to arrhythmias and neuromuscular instability. \\

    \item Non-invasive diastolic arterial blood pressure (NIDiasABP): Cuff-derived diastolic pressure; trends mirror vascular tone without an arterial line. \\

    \item Non-invasive mean arterial blood pressure (NIMAP): Cuff-based mean pressure; used when invasive monitoring is unavailable. \\

    \item Non-invasive systolic arterial blood pressure (NISysABP): Cuff-derived systolic pressure; elevations suggest hypertension or pain response. \\

    \item Serum sodium (Na): Principal extracellular cation governing osmolality; dysnatremias cause neurologic symptoms and fluid shifts. \\

    \item Partial pressure of arterial carbon dioxide (PaCO$_2$): Indicator of ventilatory status; hypercapnia implies hypoventilation, hypocapnia hyperventilation. \\

    \item Partial pressure of arterial oxygen (PaO$_2$): Measure of oxygenation efficiency; low values denote hypoxemia. \\

    \item Arterial pH: Measure of hydrogen-ion concentration; deviations from normal reflect systemic acid–base disorders. \\

    \item Platelet count (Platelets): Thrombocyte concentration essential for hemostasis; low counts increase bleeding risk, high counts thrombosis risk. \\

    \item Respiration rate (RespRate): Breaths per minute; tachypnea signals metabolic acidosis or hypoxia, bradypnea central depression. \\

    \item Hemoglobin oxygen saturation (SaO$_2$): Percentage of hemoglobin bound to oxygen; values below normal indicate significant hypoxemia. \\

    \item Invasive systolic arterial blood pressure (SysABP): Peak pressure during ventricular ejection; extremes compromise end-organ perfusion. \\

    \item Body temperature: Core temperature; fever suggests infection, hypothermia exposure or metabolic dysfunction. \\

    \item Troponin I: Cardiac-specific regulatory protein; elevation confirms myocardial injury. \\

    \item Troponin T: Isoform of cardiac troponin complex; rise parallels Troponin I in detecting myocardial necrosis. \\

    \item Urine: Hourly urine volume as a gauge of renal perfusion; oliguria signals kidney hypoperfusion or failure. \\

    \item White blood cell count (WBC): Reflects immune activity; leukocytosis suggests infection or stress, leukopenia marrow suppression or severe sepsis. \\
\end{enumerate}

\noindent\textbf{Static patient features:} Age; Gender; \emph{ICUType} – categorical code for the admitting intensive care unit (1 = Coronary Care, 2 = Cardiac Surgery Recovery, 3 = Medical ICU, 4 = Surgical ICU), capturing differences in case mix and treatment environment.

\subsection{Biomarker Subsets}
In the PhysioNet P12 task, we grouped the 36 physiological signals into one multivariate subset and 29 singleton subsets. Domain knowledge showed that only the respiratory and gas-exchange variables shared sufficiently strong, coherent dynamics to benefit from joint modeling. All other signals were physiologically diverse, so they were left as singletons to retain their unique predictive information.

\begin{enumerate}
    \item \textbf{Respiratory gas exchange and ventilation} \\
    \textit{[FiO$_2$, PaO$_2$, PaCO$_2$, SaO$_2$, RespRate, pH, MechVent]} \\
    These variables collectively describe oxygen delivery (FiO$_2$), pulmonary gas exchange efficiency (PaO$_2$, SaO$_2$), ventilatory adequacy (PaCO$_2$, RespRate, MechVent), and the resulting systemic acid–base balance (pH).  Grouping them lets the model learn the tightly coupled patterns that arise during hypoxemia, hypercapnia, mechanical ventilation adjustments, and respiratory failure—yielding a more coherent representation of a patient’s real-time respiratory status.

    \item \textbf{Singleton biomarkers} \\
    Each remaining signal represents a distinct physiological domain (hepatic, renal, hematologic, hemodynamic, neurologic, metabolic, or cardiac). Their organ-specific pathophysiology favored treating them individually, preserving granular patterns while keeping the grouping scheme simple and interpretable.
\end{enumerate}

\section{Experimental Setup and Hyperparameter Choices}
\paragraph{In-Hospital Mortality (P12) Experiments}

We trained all PhysioNet12 models for a maximum of 500 epochs using the Adam optimizer with weight decay of 1e-4. We used a ReduceLROnPlateau scheduler with a max learning rate in the \{1e-3, 1e-5\} range, min learning rate of 1e-5, factor of 0.5 and patience=20.

We trained GMAN models with batch size of range \{16, 32\}, dropout of 0.2, n\_layers in the \{3, 5\} range, hidden\_channels in the \{32, 64\} range, num\_lab\_ids\_embed in the \{5, 8\} range, num\_biom\_embed in the \{3, 5\} range, num\_units\_embed in the \{3, 5\} range.

\paragraph{Fake-News Detection (GosspiCop) Experiments}

We trained all GNAM GosspiCop models for a maximum of 400 epochs using the Adam optimizer with weight decay of 1e-4. We used a ReduceLROnPlateau scheduler with a max learning rate in the \{1e-3, 5e-5\} range, min learning rate of 1e-8, factor of 0.5 and patience=20.

\noindent We trained GMAN models with batch size of 16, dropout in \{0.0, 0.5\} range, n\_layers in the \{3, 5\} range, hidden\_channels in the \{16, 128\} range.

\noindent Random seeds were fixed for reproducibility, and results are reported across three independent runs. All models were trained on a single  NVIDIA Quadro RTX 8000 (48GB) GPU.
\end{document}